\documentclass{article}
\pdfoutput=1
\usepackage{authblk}
\usepackage[utf8]{inputenc}
\usepackage{aistats2021}
\usepackage[T1]{fontenc}    % use 8-bit T1 fonts
\usepackage{hyperref}       % hyperlinks
\usepackage{url}            % simple URL typesetting
\usepackage{booktabs}       % professional-quality tables
\usepackage{amsfonts}       % blackboard math symbols
\usepackage{nicefrac}       % compact symbols for 1/2, etc.
\usepackage{microtype}      % microtypography
\usepackage{subcaption}
\usepackage{graphicx}
\usepackage{amsmath}
\usepackage{amssymb}
\usepackage{mathtools,nccmath}
\usepackage{algorithm}
\usepackage{algorithmic}
\usepackage[compact]{titlesec}
\usepackage[round]{natbib}
\usepackage{abstract}

\newcommand{\x}{\mathbf{x}}
\newcommand{\z}{\mathbf{z}}

\newcommand{\zero}{\mathbf{0}}
\newcommand{\Id}{\mathbf{I}}

\newcommand{\R}{\mathbb{R}}
\newcommand{\Gauss}[2]{\mathcal{N}\left(#1,#2\right)}

\newcommand{\JSD}[2]{\operatorname{JSD}\left[ #1 , #2 \right]}
\newcommand{\E}[2]{\mathbb{E}_{#1}\left\{#2\right\}}
\newcommand{\K}[1]{K\left(#1\right)}
\newcommand{\G}[2]{G_{#1}\left(#2\right)}
\newcommand{\Ent}[1]{\operatorname{H}\left[ #1 \right]}

\allowdisplaybreaks
\title{\textbf{GENs: Generative Encoding Networks}}
\author{\textbf{Surojit Saha} \quad \textbf{Shireen Elhabian} \quad \textbf{Ross T. Whitaker}\\
        Scientific Computing and Imaging Institute, \\
        School of Computing, University of Utah, Salt Lake City, UT, USA\\
        surojit@cs.utah.edu, shireen@sci.utah.edu, whitaker@cs.utah.edu}
\date{}

\begin{document}

\twocolumn

\maketitle

\begin{abstract}
Mapping data from and/or onto a known family of distributions has
 become an important topic in machine learning and data analysis.
 Deep generative models (e.g., generative adversarial networks ) have
 been used effectively to match known and unknown distributions. Nonetheless, when the form of the target distribution is known, analytical methods are advantageous in providing robust results with provable properties. In this paper, we propose and analyze the use of
  nonparametric density methods to estimate the Jensen-Shannon divergence for matching unknown data distributions to known target distributions, such Gaussian or mixtures of Gaussians, in latent spaces. This analytical method has several advantages:
  better behavior when training sample quantity is low, provable
  convergence properties, and relatively few parameters, which can be derived analytically. Using the proposed method, we enforce the latent
  representation of an autoencoder to match a target distribution in a
learning framework that we call a {\em generative encoding network}. Here, we present the numerical methods; derive the expected distribution of the data in the latent space; evaluate the properties of the latent space, sample reconstruction, and generated samples; show the advantages over the adversarial counterpart; and demonstrate the application of the method in real world.
\end{abstract}

\section{Introduction}
Research in statistical modeling and deep learning has made great strides in the discovery of latent spaces in the context of nonlinear, high-dimensional, and very general transformations, such as those developed by autoencoders that employ deep neural networks \citep{tschannen2018recent}. Meanwhile, there have also been significant advances in generative models, which can systematically produce data samples from complex, high-dimensional distributions that resemble populations of training data. While encoder technologies, such as deterministic denoising \citep{vincent2008extracting} and contractive \citep{rifai2011contractive} autoencoders, can produce latent representation for samples, the structure of the population of data in that latent space is often underspecified or unconstrained, such that one cannot readily reason about that distribution or sample from them. Thus, such models often fail as generators of new samples. Generators on the other hand, such as those produced by the generative adversarial network (GAN) \citep{goodfellow2014generative}, have demonstrated impressive capabilities to produce new, very realistic samples (both qualitatively and quantitatively) from complex distributions described by training data. This is achieved, typically, by learning a transformation from a known distribution in a latent space into the high-dimensional data space.
However, the inverse mapping is often problematic (from samples into the latent space), and therefore it is often difficult to reason about the absolute or relative positions of data in that latent space.
Models that map training samples into latent spaces with well-defined properties are of significant value and interest. For instance, using such mappings, one could compute data densities in the latent space, and alleviate some of the challenges of density estimation of arbitrary distributions in high-dimensional spaces. This could be relevant for unsupervised learning methods, such as anomaly detection \citep{chalapathy2019deep}. One might also want to compare or manipulate data samples in the latent space, and thereby generate new, realistic samples in controlled ways \citep{zhu2016generative}.  These motivations, and others, have lead to more recent research that seeks to build latent spaces that have two properties (i) an approximate invertible mapping from the data space to the latent space and (ii) data in the latent space has defined statistical characteristics that lead to probabilistic reasoning and sample generation.
These developments and motivations have led to several novel architectures that attempt to unify latent-space learning and sample generation. Of particular interest for this work are the methods that structure the distribution of data in the latent space of an autoencoder. Very promising results, for instance, have been demonstrated by the adversarial autoencoders (AAE) of Makhzani et al. \citep{makhzani2016adversarial}, where a conventional autoencoder incurs an additional penalty from a discriminator in the latent space, which compares (via classification) the latent encodings of training samples to samples from a known distribution (e.g., standard normal).

In this paper, we take an alternative approach. Rather than using an adversary to compare distributions, we propose a nonadversarial framework, {\em generative encoding networks} or GENs, to enforce the latent representation of an autoencoder to a known target distribution. We penalize distributions in the latent space directly using the Jensen-Shannon divergence (JSD) of the distribution of the encoded training data and the known distribution. The density of the encoded training data is estimated with a kernel density estimator (KDE) and we compare against (match to) analytical distributions. The important insight is that known distributions, such as the standard normal, in spaces of relatively low dimension (compared to the data space) lend themselves to accurate estimates with KDEs and facilitate the automatic estimation of auxiliary or hyper-parameters, such as kernel bandwidth. GENs have several advantages. First, they avoid the complexity of solving the min-max optimization problem with competing networks, which can present challenges in choices of architecture and training strategies/parameters.  Second, we show that there are scenarios where the use of KDE-based JSD is a more accurate measure of the difference between two different distributions compared with the estimates of a corresponding adversary. Finally, the use of analytical tools allows us to reason about (or predict) the parameters and behavior of the system in ways that are not readily available with a conventional adversarial neural network.

\section{Related Work}

Directly related to GENs are deep, generative models that build low-dimensional latent spaces with known/quantifiable properties (e.g., known density such as multivariate normal) and learn a functional mapping from this latent space to the data space.
%
% VAEs 
Variational autoencoders (VAEs) \citep{kingma2014auto,rezende2014stochastic} introduce a parametric prior distribution on the latent space and learns an inference network (i.e., encoder) jointly with the generative model (i.e., decoder) to maximize a variational lower bound of the otherwise intractable marginal log-likelihood of the training data. Nonetheless, VAEs often fail to match the marginal (a.k.a. aggregate) posterior to the latent prior distribution, manifested by poor sample quality \citep{rosca2018distribution}. VAE variants, such as adversarial \citep{makhzani2016adversarial} and Wasserstein \citep{tolstikhin2017wasserstein} autoencoders, introduce structure to the aggregate posterior distribution to match the latent distribution through matching penalties (adversarial or maximum mean discrepancy, MMD, regularizers) in the latent space, which, in the limit, optimize the Jensen-Shannon divergence (JSD) of the different distributions. Other VAE variants improve matching distributions in the latent space through learning a data-driven prior distribution using normalizing flows in the low-dimensional latent space \citep{bhalodia2019dpvaes,xiao2019generative}.
To mitigate vanishing gradients in GANs \citep{arjovsky2017towards},  KernelGANs \citep{KDE_Data_Space} use a training objective that minimizes a nonparametric estimate of the JSD in the data space. However, such estimates in very high-dimensional spaces, where the data often lies, pose theoretical and computational challenges \citep{theis2015note,wu2017quantitative}, for instance, the estimation of the kernel bandwidth.
GANs, in their basic form \citep{goodfellow2014generative}, do not provide reliable mappings back into the latent space. Encoder-decoder GAN variants (e.g., BiGAN \citep{donahue2017adversarial}, ALI \citep{dumoulin2017adversarially}) simultaneously learn inference and generator networks. However, reconstructed samples from their latent representation do not preserve sample identity since the data-latent correspondence is learned through a discriminator that approximates a density ratio between their joint distribution.
Several works have introduced different GAN and autoencoder hybrids that bring the best of both worlds---generating realistic samples and approximately/softly invertible architectures \citep{AE_GANs_2017}. For instance, adversarial generator-encoder (AGE) \citep{Enc_Dec} trains an encoder and a generator in opposite directions by considering the divergence of both the encoded real data and encoded generated data to the latent prior distribution. Distributions are matched using KL divergence by empirically fitting a multivariate Gaussian with a diagonal covariance to the encoded data.

\section{Generative Encoding Networks}
\label{method_headings}

GENs augment the loss function of a conventional autoencoder with a term that penalizes the difference between distribution of training samples in the latent space (i.e., \textit{encoded} distribution) and the desired or \textit{target} distribution.  We denote the distribution of training data, in the space in which it was given (i.e., data space) as $p(\x)$ with $\x \in \R^d$, the distribution of \emph{encoded} training data in the latent space as $p_e(\z)$ with $\z \in \R^l$, and the \emph{target} distribution in the latent space as $p_t(\z)$. Let $\x_1, \ldots, \x_n$ and $\z_1, \ldots, \z_n$ be the $n-$training samples and associated latent space representations, respectively, such that $\z_i = E_\phi(\x_i)$, where $\phi$ denotes the network parameters of the encoder.

For penalizing the difference between $p_e$ and $p_t$, we use the Jensen-Shannon divergence. This penalty has the advantages of being a true metric with respect to the two distributions and having values between 0 and 1 (in the analytical case), which controls the influence of the penalty under a wide range of circumstances.  However, while the expectation can be approximated with a sample mean, it requires an estimate of the of the ratios of probabilities, which typically require density estimates and can be challenging.

Makhzani et al. \citep{makhzani2016adversarial}, in the context of GANs, note that an ideal discriminator {\em adversary} classifying samples between $p_t$ and $p_e$ with a cross entropy loss, computes the ratio of densities required for the divergence and the cross entropy is proportional to the $\JSD {p_t}{p_e}$. This is a very powerful result, but requires the design and training of a discriminator that can approximate the ideal one, with the associated hyper-parameters, architectural choices, and training strategies. In this work, we rely on the following observation: \textit{in cases where the $p_t$ has a known regular form, which it often does, we can rely on this form to compute good approximations to $p_e$ through samples, without the need of an additional neural network, or adversary.}

In principle, the target distribution, $p_{t}(\z)$ of a GEN, can be \emph{any distribution} from which we can sample, but in this paper we focus on the cases where the target distribution is Gaussian or Mixture of Gaussians, and we propose a kernel density estimate (KDE) of $p_e$ from a set of $m-$samples $\z_1, \ldots, \z_m$, such that:
\begin{equation}\label{eqn:pe}
  p_e(\z) =\frac{1}{m} \sum_{i=1}^m \K{ \frac{||\z - \z_i||}{h} } 
\end{equation}
where $h \in \R^+$ denotes the kernel bandwidth. Without loss of generality, we consider isotropic Gaussian kernels, and denote it as $\G{h}{\z - \z_i}$.

Under these conditions, the KDE-based JSD loss is as follows (leaving out constants):
\begin{align}
   &\JSD {p_e}{p_t} =  \nonumber \\
   &\E{\z' \sim p_e(\z)} {\log \left[\frac{\frac{1}{m} \sum_{i=1}^m  \G{h}{\z' - \z_i}}{p_{t}(\z') + \frac{1}{m} \sum_{i=1}^m \G{h}{\z' - \z_i}} \right]} \nonumber \\
   &+ \E{\z'' \sim \Gauss{\zero}{\Id}} { \log \left[ \frac{ p_{t}(\z'')}{p_{t}(\z'') + \frac{1}{m} \sum_{i=1}^m \G{h}{\z'' - \z_i} } \right]}
\label{eq:JSD_KDE}
\end{align}

Notice there are {\em three} sets of samples from the latent space in (\ref{eq:JSD_KDE}), denoted $\z$, $\z'$, and $\z''$.  The first set, $\z$, of size $m$, are the samples used to build the KDE of $p_e$ as in (\ref{eqn:pe}).  The second set, $\z'$ from $p_e$ are used to compute the sample mean (approximating the expectation) of the $\log$ term containing the ratio $p_e/(p_e + p_t)$. The third set,  $\z''$, are the samples from the target distribution (which can be generated analytically) that are used to compute the expectation, via the sample mean, of $p_t/(p_e + p_t)$. The choices of these three sets present some engineering decisions and how we construct derivatives and update the model.

In this work, we propose to let the samples $\z$ for the KDE represent the overall state of the encoder network $E$, with sufficient samples from the training data, and to let that set lag the updates of the autoencoder, much as we would do with an adversarial network. The expectation in the first term of (\ref{eq:JSD_KDE}), over the samples $\z'$, are used to compute the gradient, and that set comprises the {\em minibatch} (i.e., sampled from the training data), which allows us to train the encoder using stochastic gradient descent. In this case, the second term in (\ref{eq:JSD_KDE}) does not affect the update of the autoencoder, and can be ignored.

This strategy leaves us with two important decisions regarding hyper-parameters. The first is the choice of number of samples $m$ to be used in the lagging, KDE-based estimate of $p_e$ and the second is the bandwidth $h$ used in that same estimate. Of course, bandwidth selection for KDE is, in general, a very challenging problem without a completely satisfying, general solution. However, because we are quantifying differences with target distributions, where accuracy is most important when distributions are similar, we can use the target distribution itself, the dimensionality of the latent space, $l$, and the size of the sample set used in the KDE ($m$ above) to estimate the optimal bandwidth, $h_{\rm opt}$. We seek the $h_{\rm opt}$ that best differentiates two sample sets, and therefore maximizes the JSD between two sample sets from the target distribution.

Given only a dimension and a number of samples for the KDE estimate, we can estimate the optimal bandwidth by either root trapping or a fixed-point method on $h$, by evaluating the derivative of (\ref{eq:JSD_KDE}) with respect to $h$, that is $\partial \JSD{p_e}{p_t} / \partial h$. For target distribution as standard normal, Table~\ref{tab:bandwidth} shows the optimal bandwidth estimates for a variety of latent dimensions, $l$, and number of samples, $m$. As expected, the bandwidth increases with higher dimension and fewer samples.

The loss function of GENs can thus be formulated as,
\begin{align}
&\E{\x \sim p(\x)} { ||\x - D_\theta(E_\phi(\x))||_2^2 } + \\
&\E{\x \sim p(\x)} { \lambda \log \left[\frac{\frac{1}{m} \sum_{i=1}^m  \G{h}{E_\phi(\x) - \z_i}}{p_{t}({E_\phi(\x)}) + \frac{1}{m} \sum_{i=1}^m \G{h}{E_\phi(\x) - \z_i} } \right]} \nonumber
\label{eqn:gens}
\end{align}
\begin{algorithm}[H]
\caption{: \textbf{GENs training.} Minibatch stochastic gradient descent of GENs loss (3).}
\hspace*{\algorithmicindent} \textbf{Input:} Training samples $\mathcal{X} = \{\x_1, \ldots, \x_n\}$, Minibatch size $n_b$, Latent dimension $l$, Number of lagged samples for KDE $m$, Number of minibatch updates the KDE estimate should lag $k_b$, Relative scaling factor for the JSD loss $\lambda$\\
\hspace*{\algorithmicindent} \textbf{Output:} encoder parameters $\phi$ and decoder parameters $\theta$ 
\begin{algorithmic}[1]
\STATE Estimate the optimal kernel bandwidth $h_{\rm opt}$ given $(l,m)$
\STATE Initialize $\phi$ and $\theta$
\STATE Initialize lagging samples for KDE $\mathcal{Z} \subset \{E_\phi(\x_1), \ldots, E_\phi(\x_n)\}$, where $|\mathcal{Z}| = m$
\STATE Initialize minibatch index $b \gets 0$
\FOR {number of minibatch updates}
\STATE Sample a minibatch of size $n_b$ from $p(\x)$, $\mathcal{X}_b = \{\x_1, \ldots \x_{n_b}\}$
\STATE Encode minibatch samples in the latent space $\mathcal{Z}'_b = \{\z'_1, \ldots \z'_{n_b}\}$, where $\z'_i = E_\phi(\x_i)$
\STATE Update the encoder parameters, $\phi$ using stochastic gradient descent:
\STATE
\begin{equation}
    \begin{multlined}[c]
        \nabla_{\phi} \frac{1}{n_b} \sum_{i=1}^{n_b} \left\{ ||\x_i - D_\theta(E_\phi(\x_i))||_2^2 \right\} + \nonumber \\
        \lambda \nabla_{\phi} \frac{1}{n_b} \sum_{i=1}^{n_b} \left\{ \log \frac{\frac{1}{m} \sum_{i=1}^m  \G{h}{\z' - \z_i}}{p_{t}({\z'}) + \frac{1}{m} \sum_{i=1}^m \G{h}{\z' - \z_i} } \right\}
    \end{multlined}
\end{equation}
\STATE Update the decoder parameters, $\theta$ using stochastic gradient descent:
\STATE
\begin{equation}
    \begin{multlined}[c]
        \nabla_{\theta} \frac{1}{n_b} \sum_{i=1}^{n_b} ||\x_i - D_\theta(E_\phi(\x_i))||_2^2 \nonumber
    \end{multlined}
\end{equation}
\IF {$b \mod k_b$}
\STATE Update the lagging samples for KDE \{$\z_1, \ldots, \z_m$\}, where $\z_i = E_\phi(\x_i)$ using the current state of the encoder
\ENDIF
\STATE $b\gets b+1$
\ENDFOR
\end{algorithmic}
\label{alg:GEN_Alg}
\end{algorithm}
where $E_\phi$ and $D_\theta$ are the encoder and decoder with parameters $\phi$ and $\theta$, respectively. GENs simultaneously updates the encoder $E_\phi$, and the decoder $D_\theta$ to minimize the expected reconstruction loss in data space $\R^d$ and JSD in latent space $\R^l$ (using KDE of the encoded distribution). Algorithm \ref{alg:GEN_Alg} outlines GENs training.  Here, we see the relationship with several other methods. The KDE estimate of JSD acts very much like an adversary. Its single parameter, bandwidth is chosen to maximize the discrimination (also similar to an adversary) between samples from same distributions (i.e.  pessimistic assumptions). Furthermore, minimization of JSD loss (by updating the encoder parameters), set up a min-max game as in AAE---except that we know the optimal/final value of the single parameter, bandwidth, in the latent space. There is also a relationship with the VAE formulation. The aggregate latent distribution is a convolution of data samples with a kernel. The difference is that the width of that kernel is chosen specifically to aid analysis in the latent space, rather than basing it on a hypothetical noise model. Furthermore, the JSD penalty directly operates on the aggregate latent distribution, rather than individual samples, as the standard-normal prior does in the VAE. As the results section demonstrate, this results in latent distributions that are better matched to the target.

\begin{table*}[t]
    \centering
    \caption{Optimal bandwidths, $h_{\rm opt}$, estimated to maximize the JSD between two standard-normal samples sets.  Error bars are computed over 10 trials.  Notice that the bandwidth increases with increasing dimensions (vertical) or decreasing sample size (horizontal).}
    \label{tab:bandwidth}
    \begin{tabular}{|r||r|r|r|r|r|r|}
    \hline
    $l \backslash m$& 250 & 500 & 1000 & 2000 & 4000 & 8000
    \\ \hline
    2&   
    $0.46 \pm 0.08$ & $0.41 \pm 0.04$ &	$0.38 \pm 0.03$ & $0.34 \pm 0.02$ & $0.31 \pm 0.03$ & $0.28 \pm 0.01$
                                                                        \\ \hline
   5& 
    $0.65 \pm 0.03$ &	$0.59 \pm 0.02$ &	$0.55 \pm 0.02$ & $0.51 \pm 0.01$ & $0.47 \pm 0.01$ & $0.44 \pm 0.01$
    \\ \hline
    10&
    $0.77 \pm 0.02$ &	$0.73 \pm 0.01$ &	$0.69 \pm 0.01$ &	$0.66 \pm 0.01$ &	$0.63 \pm           0.01$ &	$0.6 \pm 0.01$  \\ \hline
    20 &
    $0.92 \pm 0.01$ & $0.89 \pm 0.01$ & $0.86 \pm 0.00$ & $0.82 \pm 0.01$ &	$0.80 \pm  0.00$ & $0.77 \pm 0.00 $ \\ \hline
    40 & 
    $> 1.0$ & $ > 1.0$ & $> 1.0$ & $0.98 \pm 0.00$ &$	0.95 \pm 0.00$ &$	0.94 \pm 0.00$
    \\ \hline
    \end{tabular}
  \end{table*}

  \section{Latent space distribution}
The use of analytical approach in the proposed method allows us to determine the resultant distribution in the latent space when the training of the model reaches a steady state, i.e., when the GEN successfully matches the encoded distribution, ${p_e}$, to the defined target distribution, ${p_t}$. Here, we derived the closed form expression of the resultant distribution in the latent space when the target distribution is standard normal (these results can be generalized to any multivariate Gaussian but is beyond the scope of this paper). 
When the training of GEN converges, the derivative of the JSD estimate with respect to the training samples $\z'$  approaches zero and the parameters of the encoder, $\phi$, are no longer updated.   Thus, setting ${\partial {\rm JSD}[p_e, p_t]}/{\partial \z'}$ to $0$ and rearranging the expression we get,
\begin{equation}
\begin{split}
\z'\sum_{i=1}^m  \G{h}{\z' - \z_i} &= \left(\frac{1}{h^2} \left[\sum_{i=1}^m \G{h}{\z' - \z_i}(\z' - \z_i)\right] \right)
\end{split}
\label{eq:rearrange_jsd_derivative_with_z'}
\end{equation}
The expression in (\ref{eq:rearrange_jsd_derivative_with_z'}) should hold for every $\z'$ in the latent space, and therefore constrains the configurations of 
$\z_i$'s.
Because the $\z_i$'s are stochastic (samples from held out data), we can constrain their probability distribution by taking an expectation on both sides of
(\ref{eq:rearrange_jsd_derivative_with_z'}):
\begin{align}
&\E{\z \sim p(\z_1, \z_2, ..., \z_m)} { \z'\sum_{i=1}^m  \G{h}{\z' - \z_i} } = \nonumber \\
&\E{\z \sim p(\z_1, \z_2, ..., \z_m)} {\frac{1}{h^2} \left[\sum_{i=1}^m \G{h}{\z' - \z_i}(\z' - \z_i)\right] }
\label{eq:expected_derivative}
\end{align}
The $\z_i$'s are independent (e.g., held out samples from the data space) and thus, we can marginalize the independent variables in (\ref{eq:expected_derivative}) and use the linear property of summation to get,
\begin{align}
&\z'\int_{\z_{i}} \G{h}{\z' - \z_i} p(\z_i) d\z_i =  \nonumber \\
&\frac{1}{(h^2)} \int_{\z_{i}} \G{h}{\z' - \z_i}(\z' - \z_i)  p(\z_i) d\z_i.
\label{eq:expected_derivative_Final}
\end{align}
The RHS of (\ref{eq:expected_derivative_Final}) is the expectation of $\nabla_{\z'} \G{h}{\z' - \z_i}$. By the definition of convolution and its properties we get,
\begin{equation}
\begin{split}
\z'\left[\G{h}{\z'} \circledast p(\z')\right] &= -\nabla_{\z'} \left[\G{h}{\z'} \circledast p(\z')\right] \\
Q(\z') &= \left[\G{h}{\z'} \circledast p(\z')\right] \\
\nabla_{\z'} Q(\z') &= -\z'Q(\z')
\end{split}
\label{eqn:sample_sum_jsd_derivative_with_z'}
\end{equation}
This is a partial differential equation in (\ref{eqn:sample_sum_jsd_derivative_with_z'}) with the solution (assuming appropriate boundary/limiting conditions)
\begin{equation}
Q(\z') = k * \exp \left( -||\z'||^2/2 \right) \\
\label{eqn:final_expression}
\end{equation}

Thus, $p(\z') = \G{\sigma}{\z'}$ where $\sigma^2 = 1 - h^2$. The factor $h^2$ in the variance of $p(\z')$ accounts for the bias produced by the convolution of samples used in the KDE with the kernel.   Note that for a standard-normal target distribution, the distribution of samples in the latent space as the encoder converges will degenerate (to a dirac delta function) as $h \rightarrow 1$.   This provides a practical interpretation of
Table~\ref{tab:bandwidth}, because
as $h_{\rm opt}$ in that table goes to 1, we see the lower bound on the number of samples that would be required for this method to function in any particular latent dimension.   For instance, $8000$ samples would be barely enough to estimate the JSD between samples from standard normal distributions in $l=40$ dimensions.   This limitation in samples/dimensions is a consequence of our choice of a stationary, isotropic kernel, which suggests that the method could be further improved by the application of nonstationary KDE methods.

\section{Results}
Here we present some results that demonstrate, empirically the power
of the KDE-based approach, relative to a NN adversary, to estimate
JSD. We also quantify the improvement of the latent-space
distribution on benchmark dataset while also comparing,
qualitatively and quantitatively, the generative properties of related
approaches on several datasets. Finally, we demonstrate some potential applications of
the method.

\subsection{Correlation Study}
Here, we compared the proposed KDE-based JSD estimates with the estimates from an adversarial (i.e., discriminator) neural network (NN). We create a set of experiments that control the degree of divergence between two distributions and quantify the ability of these methods to find the correlation between JSD and the parameter that controls these differences. In an ideal scenario, the computed JSD should increase with the parameter that controls the distributions' separation.

\begin{figure*}[t]
    \centering
    \includegraphics[width=0.6\textwidth]{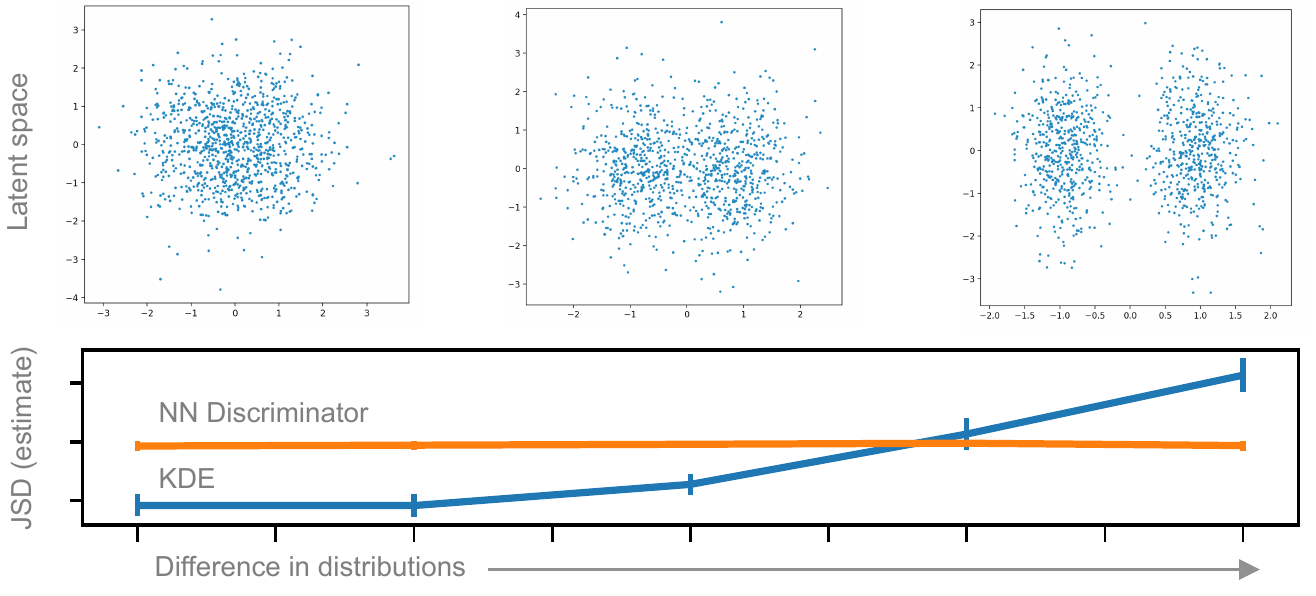}
    \includegraphics[width=0.6\textwidth]{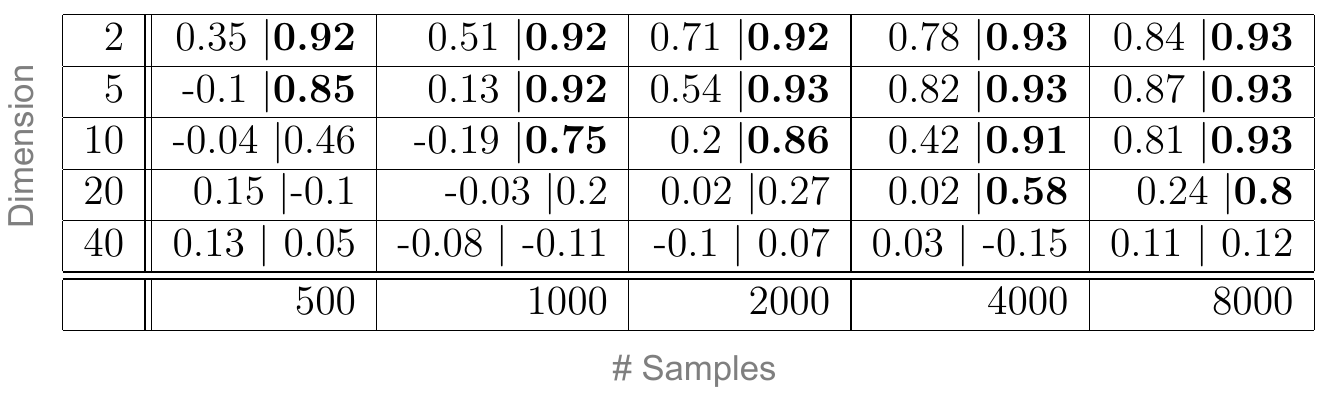}
      \caption{
      Top: Sequences of distributions, all with the same variance, that differ in controlled ways from the target (standard normal) allow us to evaluate a methods ability to quantify these differences.  A line graph with error bars shows how a nerual network (NN) and a kernel density estimator (KDE) estimate of JSD for different degrees of separation of the underlying distributions (as above) for 5 dimensions and 1000 samples.
      Bottom: Comparing distributions with varying degrees of separation (numerous trials) gives Pearson correlation coefficients for NN (left of bars) and KDE for varying dimensions and numbers of samples.  Bold indicates significant correlations that are higher than the alternative model.
      %
     % Higher dimensions require larger sample/training sizes to measure differences, with the KDE generally able to quantify differences more accurately with lower numbers of samples.
     }
    \label{fig:NDE}
\end{figure*}

For this particular set of experiments, we whiten all datasets, to
avoid trivial differences in data, and we compare all manufactured
distributions with the standard normal. We construct sequences of
distributions (proxies for a hypothetical latent distribution) by a
mixture of two standard-normal distributions, separated by distance
$s$ and subsequently whitened.  The parameter $s$ ranges from $0$ to
$6$. The top of Figure~\ref{fig:NDE} shows examples from these
distributions (in 2D) with $s = [0, 3, 6]$. This arrangement allows us
to study the behavior of these different methods over multiple trials,
different dimensions, $l$, and different training sample sizes,
$m$.

The single parameter, the bandwidth, for the GEN is computed using
the optimization described in Section~\ref{method_headings}.
The NN-discriminator has three hidden layers, with the
number of units at each layer proportional to $l$, and the first layer
being the widest, with $10 \times l$
units, and it was regularized using an early stopping criteria which
evaluated held-out data. 
%The network was regularized using early stopping by evaluating a validation data (of size $n_v$) and terminating training at the point where the loss function evaluated on the validation data rises. In all experiments, the network achieved this condition within $5-10$ epochs.

The middle panel of Figure~\ref{fig:NDE} shows a graph of a particular
experiment, $l = 5$ and $m=1000$,
% we need to verify the number of samples
the estimated JSD as a function of $s$, with error bars showing over 20
trials.  The KDE shows an upward trend, while the NN was unable to
detect the progressive increase in differences. These slopes and the
corresponding error bars are important, because the derivatives of the
JSD estimates are what drive the structure of the latent space in this
autoencoder context.

Using this method we can quantify the slope/noise relationship for
each method as a function of dimension and sample size.
The lower panel of Figure~\ref{fig:NDE}
reports the Pearson correlations over a range of dimensions and sample
sizes.  In each cell,
the NN|KDE are on the left|right, respectively. Numbers in bold
indicate higher correlations that are statistically significant. These
results demonstrate that the estimation of JSD using NN is strongly
influenced by sample size and data dimension. In many cases, the KDE
is able to quantify correlation where the NN does not.

\subsection{Evaluating structure in learned latent spaces}
Here we quantify how well-formed the Gaussian structure is in the
latent space after the autoencoder is optimized. We rely on the fact
that for a given covariance, $\mathbf{\Sigma}$, the Gaussian has the
maximum entropy among all distributions with the same
covariance. Thus, we can compute the entropy, $\Ent{p_e}$, of the
whitened latent distributions from various learned models to evaluate
the {\em normality} of the encoded distribution for different methods. Evaluating entropy relies on estimates of density. For this, we also
use a KDE, as defined in (\ref{eqn:pe}). For a certain latent
dimension $l$, and sample size $m$, the optimal bandwidth
$h_{\rm opt}$ can be determined by minimizing the entropy of the KDE
estimate of a distribution given samples from it (maximum log
likelihood). This is in contrast to estimating the bandwidth using JSD
that best differentiates two distributions, encoded and target.
To find $h_{\rm opt}$ that minimizes the entropy of a set of latent space samples, we differentiate $\Ent{ p_e}$ with respect to $h$, set it equal to zero, and solve for $h_{\rm opt}$ with a fixed point strategy. We have, 
\begin{equation}\label{Average_NLL}
\Ent{ p_e(\z)} \approx -\frac{1}{m} \sum_i \log \frac{1}{m-1} \sum_{j \ne i} \K{ \frac{||\z_j - \z_i||}{h} },
\end{equation}
%which gives the fixed point update:
%\begin{equation}\label{fixedpoint}
%h^2  \leftarrow \frac{1}{l m }\sum_i  \frac{\sum_{j \ne i} {||\z_j-\z_i||_2^2} \G{h}{\z_j - \z_i}}{\sum_{j \ne i} \G{h}{\z_j - \z_i}}.
%\end{equation}

\begin{table*}
    \centering
    \caption{Mean and standard deviation of entropy of the latent representations generated with AAE \citep{makhzani2016adversarial}, VAE \citep{kingma2014auto} and GEN using the \textit{FC-layers} architecture over MNIST data set with latent dimension of $l=8$.}     \label{tab:Entropy_Latent_Space_1}
    \resizebox{\textwidth}{!}{%
    \begin{tabular}{|l|c|c|c|c|}\hline
    $Method \backslash Samples$ & 1000 & 2000 & 5000 & 10000 \\ \hline
    AAE & $6.552 \pm 1.8164$ & $8.408 \pm 0.5280$ & $8.742 \pm 0.4209$ & $9.862 \pm 1.1321$ \\
    VAE & $10.478 \pm 0.1647$ & $10.483 \pm 0.2013$ & $10.347 \pm 0.0769$ & $10.130 \pm 0.1516$ \\
    GEN (proposed) & $11.681 \pm 0.0231$ & $11.622 \pm 0.0170$ & $11.497 \pm 0.0068$ & $11.464 \pm 0.0036$ \\
    Standard Normal & $11.668 \pm 0.0552$ & $11.621 \pm 0.0597$ & $11.575 \pm 0.0200$ & $11.538 \pm 0.0162$ \\\hline
    \end{tabular}}
\end{table*}

We evaluate the entropy of the latent distributions for the proposed
GEN, AAE\citep{makhzani2016adversarial} and VAE \citep{kingma2014auto}
for varying training sample sizes.   If the latent
representation closely matches the targeted, standard-normal distribution, then it
its entropy will match that of standard normal for that dimension,
which has analytical expression. For a
fair comparison, the architecture of the $E_\phi$ and $D_\theta$ used
is same for all the methods. For this experiment, we have used only
fully connected (FC) layers, and henceforth we refer to this
architecture as \textit{FC-layers}. For statistical evaluation, we
have trained $5$ different models of the network for each method for
different settings.
%Before evaluation of entropy, the latent
%representations for all models are whitened to account for systematic
%biases in covariance. 
This experiment is conducted on MNIST data set
with latent dimension of $l=8$. Table
\ref{tab:Entropy_Latent_Space_1}, shows that the entropy of GEN
is consistently higher than that of AAE and VAE for all the sample
sizes and almost equal to the entropy estimated for samples drawn from
standard-normal distribution using the same numerical methods. 
The true entropy of the standard normal distribution with $l=8$ is $11.35$, very close to
our numerical estimates and that of the GEN. The observation for VAE is consistent with \citep{rosca2018distribution}.
For the data reported in Table
\ref{tab:Entropy_Latent_Space_1}, the reconstruction losses of GEN,
AAE\citep{makhzani2016adversarial} and VAE\citep{kingma2014auto} for all
sample sizes were observed to be approximately $0.02$, close to the loss of
unconstrained autoencoder, ruling that out as a confounding factor in
the reported differences.

\subsection{Training of GENs on bench-mark data}
\begin{figure*}[htb]
    \centering
    \includegraphics[width=0.95\textwidth]{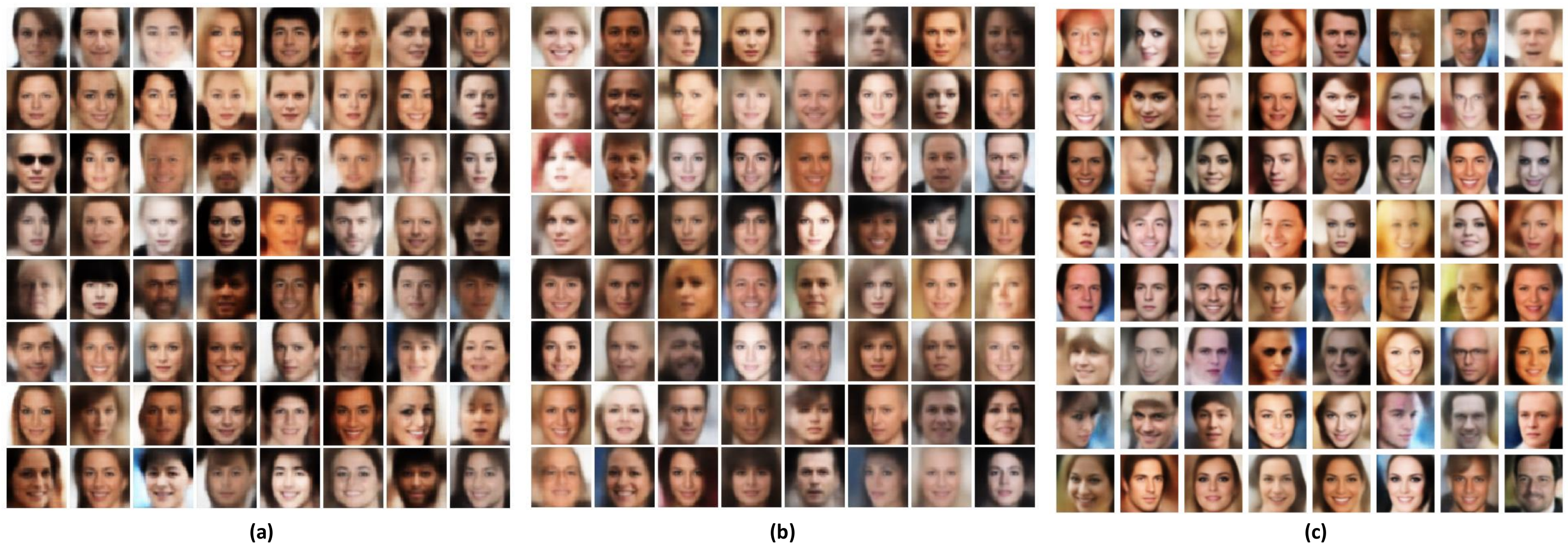}
    \caption{Generated images for CelebA dataset in $l=40$ dimensional latent space for different methods, (a) AAE \citep{makhzani2016adversarial} (b) VAE \citep{kingma2014auto} and (c) GEN (proposed)}
    \label{fig:CelebA_Generated_Images}
\end{figure*}

We examine qualitative results for several benchmark datasets: MNIST,
CelebA and SVHN, with a standard-normal target. 
For training GEN, we use a subset of the training data for the
lagging samples of the KDE-estimate. This size, $m$, 
is set to $20K$ for CelebA and $10K$ for MNIST and SVHN. The
latent dimensions, $l$, are $8, 25$ and $40$ for MNIST, SVHN and CelebA
respectively. The relative scaling factor $\lambda$, for JSD loss was
in the range $[0.01, 0.005]$ for different scenarios. In all our
experiments, we have used Adam optimizer (learning rate: $2e^{-04}$,
$\beta_1$: $0.5$). Number of minibatch updates the KDE estimate should
lag, $k_b$, is set to $10$. The FID score of GEN and other competing
methods for different datasets are reported in Table
\ref{tab:FID_Score}. Qualitative results of the different methods
trained over CelebA dataset are shown in Figure
\ref{fig:CelebA_Generated_Images}. Quantitative results show that
the GEN generates marginally better data-space examples than AAE and
VAE. This can be confirmed, qualitatively, by the generated samples in Figure~\ref{fig:CelebA_Generated_Images}.

\begin{table}[!htb]
    \centering
    \caption{FID score of competing methods for different datasets (lower is better)}
    \label{tab:FID_Score}
    \resizebox{\columnwidth}{!}{
    \begin{tabular}{|l|c|c|c|}\hline
     & AAE & VAE & GEN \\ \hline
     SVHN & $56.08 \pm 0.35$ & $ 55.21 \pm 0.24$ & $\mathbf{45.52 \pm 0.33}$ \\
     CelebA & $52.65 \pm 0.06$ & $55.51 \pm 0.25$ & $\mathbf{43.65 \pm 0.09}$\\\hline
    \end{tabular}}
\end{table}

\begin{figure}[htb]
    \centering
    \includegraphics[width=0.41\textwidth]{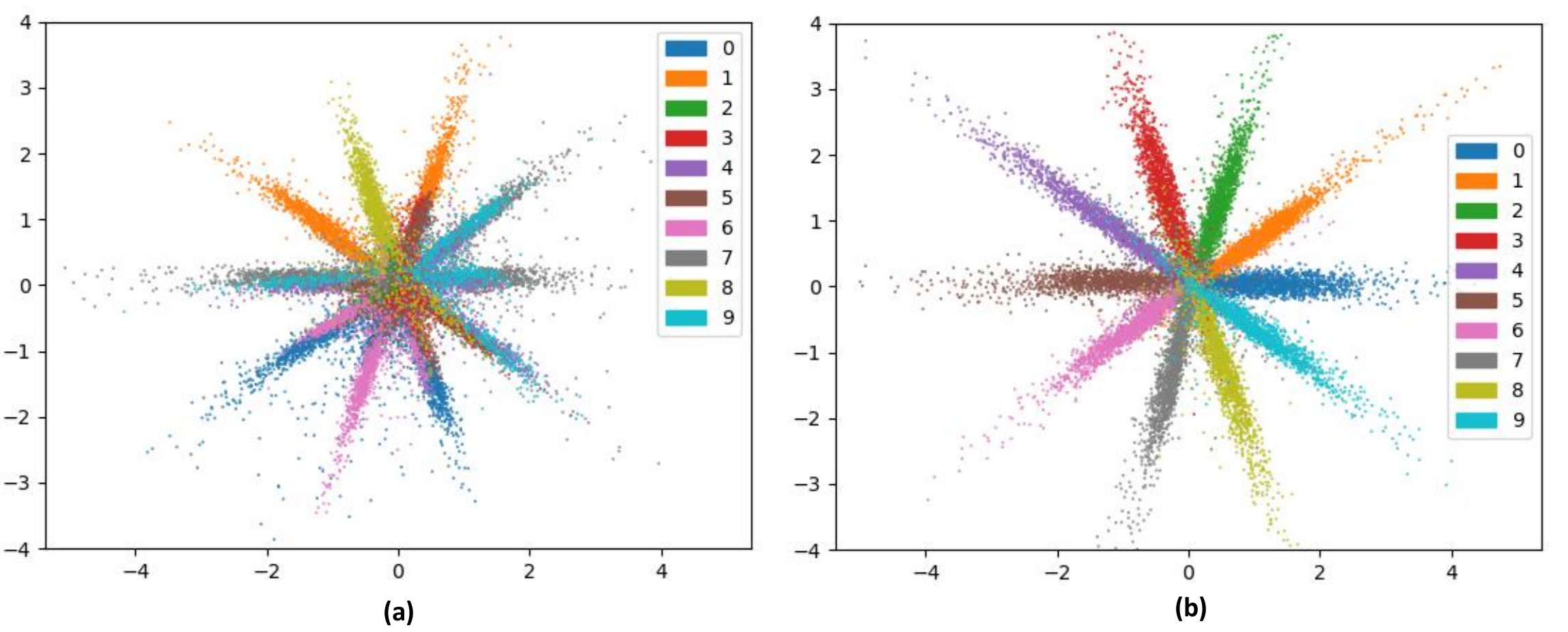}
    \caption{Mapping of MNIST to 10 2D Gaussians in the latent space (a) without any supervision (b) with labels for $10K$ samples.}
    \label{fig:Semi-supervised}
\end{figure}

\subsection{Semi-supervised learning}
In this experiment, we model the distribution in the latent space as
mixture of Gaussians (MoG) distributions where each mode in the
mixture corresponds to a category of input data, similar to the
results in \citep{makhzani2016adversarial}. To map input data to
a specific mode of the distribution (depending on its category), we
have used the labels of the data for a small set of examples. The idea
is to leverage these partially labeled examples to map unlabeled data to
its corresponding mode. This experiment was designed on MNIST dataset
for the target distribution as the mixture of 10 Gaussians where each
mode represents different digit $[0-9]$. For training, we have used
$10K$ labeled samples, and $40K$ unlabeled. For the KDE estimate
of the encoded distribution, $p_{e}$, we have used $m=10K$ samples,
separate from the training data. The optimum bandwidth for KDE,
$h_{\rm opt}$, is determined by using samples from the target
distribution (MoG) such that it maximizes the JSD between two sample
sets from the target distribution. Projection of the held out data in
the latent space using a trained model is shown in Figure
\ref{fig:Semi-supervised}, demonstrating that the GEN maps the test
inputs to the correct mode. 

\subsection{Novelty detection}
\begin{figure}[h]
    \centering
    \includegraphics[width=0.43\textwidth]{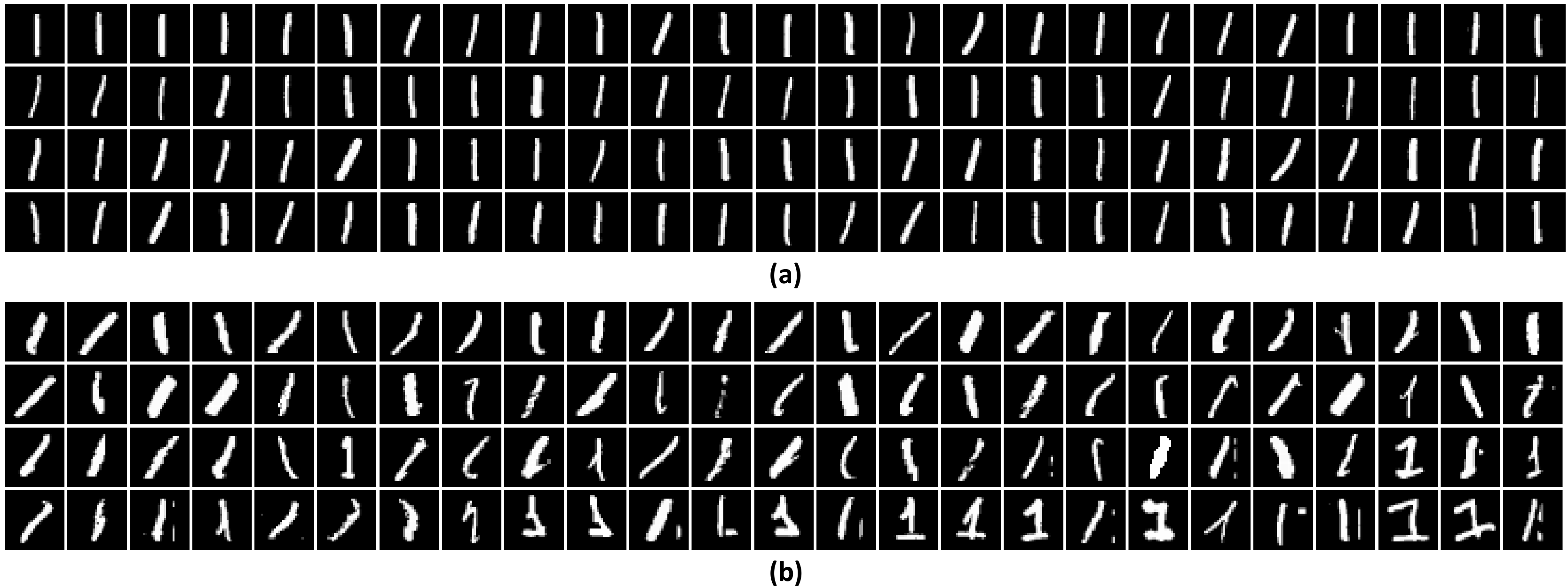}
    \caption{Novelty detection on the held out data of MNIST for digit $1$ (a) $100$ samples with low outlier score aka inliers and (b) $100$ samples having high outlier score, identified as outliers to the distribution}
    \label{fig:Outlier-Detection}
\end{figure}

Novelty detection is the method of detecting the rare events or
outliers relative to the prevalent samples (aka inliers). Most of the
methods for outlier detection, using the encoder-decoder architecture,
relies on the reconstruction loss \citep{Reconstruction_Outlier_2015}
or the employs a one-class classifier \citep{One_Class2018} to predict
outliers. Here, we detect anomaly using the estimated
probability in the data space, which is similar in philosophy to
\citep{Novelty_Detection_AAE}, but we present a very different
mathematical formulation. Strictly invertible networks
\citep{dinh2017density} can estimate data density using latent space
density and transformation Jacobians. In the encoder-decoder architecture of
an autoencoder, we must account for the lower dimension of the model in
the data space. Thus, we construct the probability in the data space using
the in-manifold probability (manifold of dimension $l$ projected from
the latent to data space, parameterized by the probability in the
latent space and Jacobian of the mapping) and off-manifold probability
(projection of sample on the manifold in data space). Probability in
the data space is thus computed as follows:
\begin{align}
   &p(\x) = p(\z'){\left\lvert \det{\frac{\partial \x'}{\partial \z'}} \right\rvert}^{-1} p(\x|\x') \\ %p(\x-\x')
   &\text{Where, } \z' = E_\phi(\x) \text{ and } \x' = D_\theta(E_\phi(\x)) \nonumber \\
   & p(\z'){\left\lvert \det{\frac{\partial \x'}{\partial \z'} } \right\rvert}^{-1} \text{ is the in-manifold probability} \nonumber \\
   & p(\x|\x') = \Gauss{\x;\x'}{\sigma^2\Id} \text{ is the off-manifold probability} \nonumber % p(\x-\x') = \Gauss{\zero}{\sigma^2\Id}
\label{eq:Oulier_Detection}
\end{align}
For the effects of change of coordinates, we use the first fundamental
form to compute $\det{\frac{\partial \x'}{\partial \z'}}$ as,
\begin{equation}
    \text{Let }
    \mathbf{A} = \frac{\partial \x'}{\partial \z'}
    \text{ and }
    \mathbf{B} = \mathbf{A}^T\mathbf{A}
    \text{ then, }
    \det{\mathbf{A}} = \sqrt{\lvert \det{\mathbf{B}} \rvert}
\end{equation}
This outlier detection method is applied over the MNIST dataset
and the results are shown in Figure~\ref{fig:Outlier-Detection}. In
this experiment, outlier prediction is done for individual category of
digits in the held out data. The outlier score for a test sample is
computed as the negative log of the probability in the data space. 
We have shown the 100 highest and lowest probability test samples, sorted by the outlier
scores, in Figure~\ref{fig:Outlier-Detection}. The first $100$ samples
with low score are considered inliers to the distribution, while the
last $100$ are identified as outliers, which possibly belong to the
low probability region.

\section{Discussion}
The proposed GEN method is very promising in its ability to generate
latent space representations that adhere to target distributions and
associated generative models. The method has fewer parameters than
alternatives and produces results that compare favorably with previous
methods in terms of both the latent space structure and the generated
data. One important remaining challenge is dimensionality.  While we
report results for $l=40$, the method starts to break down above
$l=50$, where optimal bandwidths are too high for virtually any number
of samples and KDE-estimates with stationary kernels become
impractical. This is, of course, the classic
curse of dimensionality, which any method that attempts to
compute statistics in such spaces faces. However, the method could almost
certainly be pushed into somewhat high dimensions using more
sophisticated KDE methods---one area of further investigation.

\bibliographystyle{abbrvnat}
\setcitestyle{authoryear,open={((},close={))}}
\bibliography{references}

\begin{thebibliography}{25}
\providecommand{\natexlab}[1]{#1}
\providecommand{\url}[1]{\texttt{#1}}
\expandafter\ifx\csname urlstyle\endcsname\relax
  \providecommand{\doi}[1]{doi: #1}\else
  \providecommand{\doi}{doi: \begingroup \urlstyle{rm}\Url}\fi

\bibitem[Arjovsky and Bottou(2017)]{arjovsky2017towards}
M.~Arjovsky and L.~Bottou.
\newblock Towards principled methods for training generative adversarial
  networks.
\newblock \emph{arXiv preprint arXiv:1701.04862}, 2017.

\bibitem[Bhalodia et~al.(2019)Bhalodia, Lee, and Elhabian]{bhalodia2019dpvaes}
R.~Bhalodia, I.~Lee, and S.~Elhabian.
\newblock dpvaes: Fixing sample generation for regularized vaes.
\newblock \emph{arXiv preprint arXiv:1911.10506}, 2019.

\bibitem[Chalapathy and Chawla(2019)]{chalapathy2019deep}
R.~Chalapathy and S.~Chawla.
\newblock Deep learning for anomaly detection: A survey.
\newblock \emph{arXiv preprint arXiv:1901.03407}, 2019.

\bibitem[Dinh et~al.(2017)Dinh, Sohl-Dickstein, and Bengio]{dinh2017density}
L.~Dinh, J.~Sohl-Dickstein, and S.~Bengio.
\newblock Density estimation using real nvp.
\newblock \emph{International Conference on Learning Representations (ICLR)},
  2017.

\bibitem[Donahue et~al.(2017)Donahue, Kr{\"a}henb{\"u}hl, and
  Darrell]{donahue2017adversarial}
J.~Donahue, P.~Kr{\"a}henb{\"u}hl, and T.~Darrell.
\newblock Adversarial feature learning.
\newblock \emph{International Conference on Learning Features (ICLR)}, 2017.

\bibitem[Dumoulin et~al.(2017)Dumoulin, Belghazi, Poole, Mastropietro, Lamb,
  Arjovsky, and Courville]{dumoulin2017adversarially}
V.~Dumoulin, I.~Belghazi, B.~Poole, O.~Mastropietro, A.~Lamb, M.~Arjovsky, and
  A.~Courville.
\newblock Adversarially learned inference.
\newblock \emph{International Conference on Learning Features (ICLR)}, 2017.

\bibitem[Goodfellow et~al.(2014)Goodfellow, Pouget-Abadie, Mirza, Xu,
  Warde-Farley, Ozair, Courville, and Bengio]{goodfellow2014generative}
I.~Goodfellow, J.~Pouget-Abadie, M.~Mirza, B.~Xu, D.~Warde-Farley, S.~Ozair,
  A.~Courville, and Y.~Bengio.
\newblock Generative adversarial nets.
\newblock In \emph{Advances in neural information processing systems}, pages
  2672--2680, 2014.

\bibitem[Kingma and Welling(2014)]{kingma2014auto}
D.~P. Kingma and M.~Welling.
\newblock Auto-encoding variational bayes.
\newblock \emph{International Conference on Learning Representations}, 2014.

\bibitem[Makhzani et~al.(2016)Makhzani, Shlens, Jaitly, Goodfellow, and
  Frey]{makhzani2016adversarial}
A.~Makhzani, J.~Shlens, N.~Jaitly, I.~Goodfellow, and B.~Frey.
\newblock Adversarial autoencoders.
\newblock In \emph{International Conference on Learning Representations}, 2016.

\bibitem[Pidhorskyi et~al.(2018)Pidhorskyi, Almohsen, and
  Doretto]{Novelty_Detection_AAE}
S.~Pidhorskyi, R.~Almohsen, and G.~Doretto.
\newblock Generative probabilistic novelty detection with adversarial
  autoencoders.
\newblock In \emph{NIPS}, 2018.

\bibitem[Rezende et~al.(2014)Rezende, Mohamed, and
  Wierstra]{rezende2014stochastic}
D.~J. Rezende, S.~Mohamed, and D.~Wierstra.
\newblock Stochastic backpropagation and approximate inference in deep
  generative models.
\newblock In \emph{International Conference on Machine Learning}, pages
  1278--1286, 2014.

\bibitem[Rifai et~al.(2011)Rifai, Vincent, Muller, Glorot, and
  Bengio]{rifai2011contractive}
S.~Rifai, P.~Vincent, X.~Muller, X.~Glorot, and Y.~Bengio.
\newblock Contractive auto-encoders: Explicit invariance during feature
  extraction.
\newblock In \emph{In International Conference on Machine Learning}. Citeseer,
  2011.

\bibitem[Rosca et~al.(2017)Rosca, Lakshminarayanan, Warde-Farley, and
  Mohamed]{AE_GANs_2017}
M.~Rosca, B.~Lakshminarayanan, D.~Warde-Farley, and S.~Mohamed.
\newblock Variational approaches for auto-encoding generative adversarial
  networks.
\newblock In \emph{arXiv preprint arXiv:1706.04987}, 2017.

\bibitem[Rosca et~al.(2018)Rosca, Lakshminarayanan, and
  Mohamed]{rosca2018distribution}
M.~Rosca, B.~Lakshminarayanan, and S.~Mohamed.
\newblock Distribution matching in variational inference.
\newblock \emph{arXiv preprint arXiv:1802.06847}, 2018.

\bibitem[Sabokrou et~al.(2018)Sabokrou, Khalooei, Fathy, and
  Adeli]{One_Class2018}
M.~Sabokrou, M.~Khalooei, M.~Fathy, and E.~Adeli.
\newblock Adversarially learned one-class classifier for novelty detection.
\newblock In \emph{CVPR}, 2018.

\bibitem[Sinn and Rawat(2018)]{KDE_Data_Space}
M.~Sinn and A.~Rawat.
\newblock Non-parametric estimation of jensen-shannon divergence in generative
  adversarial network training.
\newblock In \emph{AISTATS}, 2018.

\bibitem[Theis et~al.(2015)Theis, Oord, and Bethge]{theis2015note}
L.~Theis, A.~v.~d. Oord, and M.~Bethge.
\newblock A note on the evaluation of generative models.
\newblock \emph{arXiv preprint arXiv:1511.01844}, 2015.

\bibitem[Tolstikhin et~al.(2017)Tolstikhin, Bousquet, Gelly, and
  Schoelkopf]{tolstikhin2017wasserstein}
I.~Tolstikhin, O.~Bousquet, S.~Gelly, and B.~Schoelkopf.
\newblock Wasserstein auto-encoders.
\newblock \emph{arXiv preprint arXiv:1711.01558}, 2017.

\bibitem[Tschannen et~al.(2018)Tschannen, Bachem, and
  Lucic]{tschannen2018recent}
M.~Tschannen, O.~Bachem, and M.~Lucic.
\newblock Recent advances in autoencoder-based representation learning.
\newblock \emph{arXiv preprint arXiv:1812.05069}, 2018.

\bibitem[Ulyanov et~al.(2018)Ulyanov, Vedaldi, and Lempitsky]{Enc_Dec}
D.~Ulyanov, A.~Vedaldi, and V.~Lempitsky.
\newblock It takes (only) two: Adversarial generator-encoder networks.
\newblock 2018.

\bibitem[Vincent et~al.(2008)Vincent, Larochelle, Bengio, and
  Manzagol]{vincent2008extracting}
P.~Vincent, H.~Larochelle, Y.~Bengio, and P.-A. Manzagol.
\newblock Extracting and composing robust features with denoising autoencoders.
\newblock In \emph{Proceedings of the 25th international conference on Machine
  learning}, pages 1096--1103, 2008.

\bibitem[Wu et~al.(2017)Wu, Burda, Salakhutdinov, and
  Grosse]{wu2017quantitative}
Y.~Wu, Y.~Burda, R.~Salakhutdinov, and R.~Grosse.
\newblock On the quantitative analysis of decoder-based generative models.
\newblock \emph{International Conference on Learning Representations (ICLR)},
  2017.

\bibitem[Xia et~al.(2015)Xia, Cao, Wen, Hua, and
  Sun]{Reconstruction_Outlier_2015}
Y.~Xia, X.~Cao, F.~Wen, G.~Hua, and J.~Sun.
\newblock Learning discriminative reconstructions for unsupervised outlier
  removal.
\newblock In \emph{ICCV}, 2015.

\bibitem[Xiao et~al.(2019)Xiao, Yan, Chen, and Amit]{xiao2019generative}
Z.~Xiao, Q.~Yan, Y.~Chen, and Y.~Amit.
\newblock Generative latent flow: A framework for non-adversarial image
  generation.
\newblock \emph{arXiv preprint arXiv:1905.10485}, 2019.

\bibitem[Zhu et~al.(2016)Zhu, Kr{\"a}henb{\"u}hl, Shechtman, and
  Efros]{zhu2016generative}
J.-Y. Zhu, P.~Kr{\"a}henb{\"u}hl, E.~Shechtman, and A.~A. Efros.
\newblock Generative visual manipulation on the natural image manifold.
\newblock In \emph{European Conference on Computer Vision}, pages 597--613.
  Springer, 2016.

\end{thebibliography}


\begin{thebibliography}{2}
\providecommand{\natexlab}[1]{#1}
\providecommand{\url}[1]{\texttt{#1}}
\expandafter\ifx\csname urlstyle\endcsname\relax
  \providecommand{\doi}[1]{doi: #1}\else
  \providecommand{\doi}{doi: \begingroup \urlstyle{rm}\Url}\fi

\bibitem[Kingma and Welling(2014)]{kingma2014auto}
D.~P. Kingma and M.~Welling.
\newblock Auto-encoding variational bayes.
\newblock \emph{International Conference on Learning Representations}, 2014.

\bibitem[Makhzani et~al.(2016)Makhzani, Shlens, Jaitly, Goodfellow, and
  Frey]{makhzani2016adversarial}
A.~Makhzani, J.~Shlens, N.~Jaitly, I.~Goodfellow, and B.~Frey.
\newblock Adversarial autoencoders.
\newblock In \emph{International Conference on Learning Representations}, 2016.

\end{thebibliography}

\end{document}

\onecolumn
\maketitle

\section{Bandwidth estimation}
\subsection{Optimal bandwidth estimation for GENs}

The KDE-based JSD loss, $\JSD {p_e}{p_t}$, between the encoded distribution, ${p_e}$, and target distribution, ${p_t}$, as defined in the main paper (leaving out the constants), is as follows:

\begin{align}
   &\JSD {p_e}{p_t} =  \nonumber \\
   &\E{\z' \sim p_e(\z)} {\log \left[\frac{\frac{1}{m} \sum_{i=1}^m  \G{h}{\z' - \z_i}}{p_{t}(\z') + \frac{1}{m} \sum_{i=1}^m \G{h}{\z' - \z_i}} \right]} \nonumber \\
   &+ \E{\z'' \sim p_{t}(\z)} { \log \left[ \frac{ p_{t}(\z'')}{p_{t}(\z'') + \frac{1}{m} \sum_{i=1}^m \G{h}{\z'' - \z_i} } \right]}
\label{eq:JSD_KDE}
\end{align}

We require the derivative of $\JSD {p_e}{p_t}$, with respect to bandwidth, $h$, to determine
the optimal bandwidth, $h_{\rm opt}$, for a given latent dimension, $l$ and KDE sample size, $m$. The derivative of $\JSD {p_e}{p_t}$, with respect to bandwidth, $h$, is as follows:
\begin{equation}
\begin{split}
\frac{\partial {\rm JSD}[p_e, p_t]}{\partial h} &= \mathbb{E}_{\z' \sim p_e(\z)} \left(\frac{1}{h} \left[\frac{1}{ \sum_{i=1}^m  \G{h}{\z' - \z_i}}\right]  \left[\sum_{i=1}^m \G{h}{\z' - \z_i}\frac{||\z' - \z_i||_2^2}{h^2}\right] - \frac{l}{h} \right)\\
&- \mathbb{E}_{\z' \sim p_e(\z)} \left( \frac{1}{mh}\left[\frac{1}{{p_t}({\z'}) + \frac{1}{m} \sum_{i=1}^m  \G{h}{\z' - \z_i}} \right]\left[ \sum_{i=1}^m \G{h}{\z' - \z_i}\frac{||\z' - \z_i||_2^2}{h^2}\right] \right) \\
&+ \mathbb{E}_{\z' \sim p_e(\z)} \left( \left[\frac{\frac{1}{m} \sum_{i=1}^m \G{h}{\z' - \z_i}}{{p_t}({\z'}) + \frac{1}{m} \sum_{i=1}^m \G{h}{\z' - \z_i}} \right] \frac{l}{h} \right) \\
&- \mathbb{E}_{\z'' \sim {p_t}({\z})} \left( \frac{1}{mh}\left[\frac{1}{{p_t}({\z''}) + \frac{1}{m} \sum_{i=1}^m  \G{h}{\z'' - \z_i}} \right]\left[ \sum_{i=1}^m \G{h}{\z'' - \z_i}\frac{||\z'' - \z_i||_2^2}{h^2}\right] \right) \\
&+ \mathbb{E}_{\z'' \sim {p_t}({\z})} \left( \left[\frac{\frac{1}{m} \sum_{i=1}^m \G{h}{\z'' - \z_i}}{{p_t}({\z''}) + \frac{1}{m} \sum_{i=1}^m \G{h}{\z'' - \z_i}} \right] \frac{l}{h} \right)
\end{split}
\label{eqn:jsd_derivative}
\end{equation}

The optimal bandwidth, $h_{\rm opt}$, can be determined by either root trapping or fixed-point method using the derivative of $\JSD {p_e}{p_t}$ with respect to bandwidth, $h$. In our implementation, we have used root trapping method to determine $h_{\rm opt}$. The solution for fixed-point update is derived in (\ref{eqn:fixed_update}). Setting the derivative, $\frac{\partial {\rm JSD}}{\partial h}$ in (\ref{eqn:jsd_derivative}), to zero and multiplying both sides by $h^3$, we get the fixed-point update as follows:
%\begin{comment}
\begin{equation}
\begin{split}
h^2 &\leftarrow F^{-1} \mathbb{E}_{\z' \sim p_e(\z)} \left( \left[\frac{1}{ \sum_{i=1}^m  \G{h}{\z' - \z_i}}\right]  \left[\sum_{i=1}^m \G{h}{\z' - \z_i}||\z' - \z_i||_2^2\right] \right)\\
&- F^{-1} \mathbb{E}_{\z' \sim p_e(\z)} \left( \frac{1}{m}\left[\frac{1}{{p_t}({\z'}) + \frac{1}{m} \sum_{i=1}^m  \G{h}{\z' - \z_i}} \right]\left[ \sum_{i=1}^m \G{h}{\z' - \z_i}||\z' - \z_i||_2^2\right] \right) \\
&- F^{-1} \mathbb{E}_{\z'' \sim {p_t}({\z})} \left( \frac{1}{m}\left[\frac{1}{{p_t}({\z''}) + \frac{1}{m} \sum_{i=1}^m  \G{h}{\z'' - \z_i}} \right]\left[ \sum_{i=1}^m \G{h}{\z'' - \z_i}||\z'' - \z_i||_2^2\right] \right)  \nonumber
\end{split}
\label{eqn:fixed_update}
\end{equation}

\begin{equation}
\begin{split}
\text{where, }F &= l \\
&- \mathbb{E}_{\z' \sim p_e(\z)} \left( \left[\frac{\frac{1}{m} \sum_{i=1}^m \G{h}{\z' - \z_i}}{{p_t}({\z'}) + \frac{1}{m} \sum_{i=1}^m \G{h}{\z' - \z_i}} \right] l \right) \\
&- \mathbb{E}_{\z'' \sim {p_t}({\z})} \left( \left[\frac{\frac{1}{m} \sum_{i=1}^m \G{h}{\z'' - \z_i}}{{p_t}({\z''}) + \frac{1}{m} \sum_{i=1}^m \G{h}{\z'' - \z_i}} \right] l \right)
\end{split}
\label{eqn:lhs_factor}
\end{equation}
%\end{comment}

\subsection{Optimal bandwidth estimation for entropy minimization}
In the main paper, for target distribution, ${p_t}=\Gauss{\zero}{\Id}$, we study how well-formed the Gaussian structure is in the latent space for different methods, after the autoencoder is optimized, by evaluating the entropy of the encoded distribution, ${p_e}$. The KDE based entropy using Gaussian kernels is defined as follows:
\begin{equation}\label{Average_NLL}
\Ent{ p_e(\z)} \approx -\frac{1}{m} \sum_i \log \frac{1}{m-1} \sum_{j \ne i} \G{h}{\z - \z_i},
\end{equation}
In this experiment, the optimum bandwidth, $h_{\rm opt}$, is determined using the derivative of $\Ent{ p_e}$ with respect to $h$ and solve for $h_{\rm opt}$ with a fixed point strategy, which gives the fixed point update as,
\begin{equation}\label{fixedpoint}
h^2  \leftarrow \frac{1}{l m }\sum_i  \frac{\sum_{j \ne i} {||\z_j-\z_i||_2^2} \G{h}{\z_j - \z_i}}{\sum_{j \ne i} \G{h}{\z_j - \z_i}}.
\end{equation}

\section{Results on bench-mark data}

\subsection{Reconstruction of test samples}
\begin{figure}
    \centering
    \includegraphics[width=1.0\textwidth]{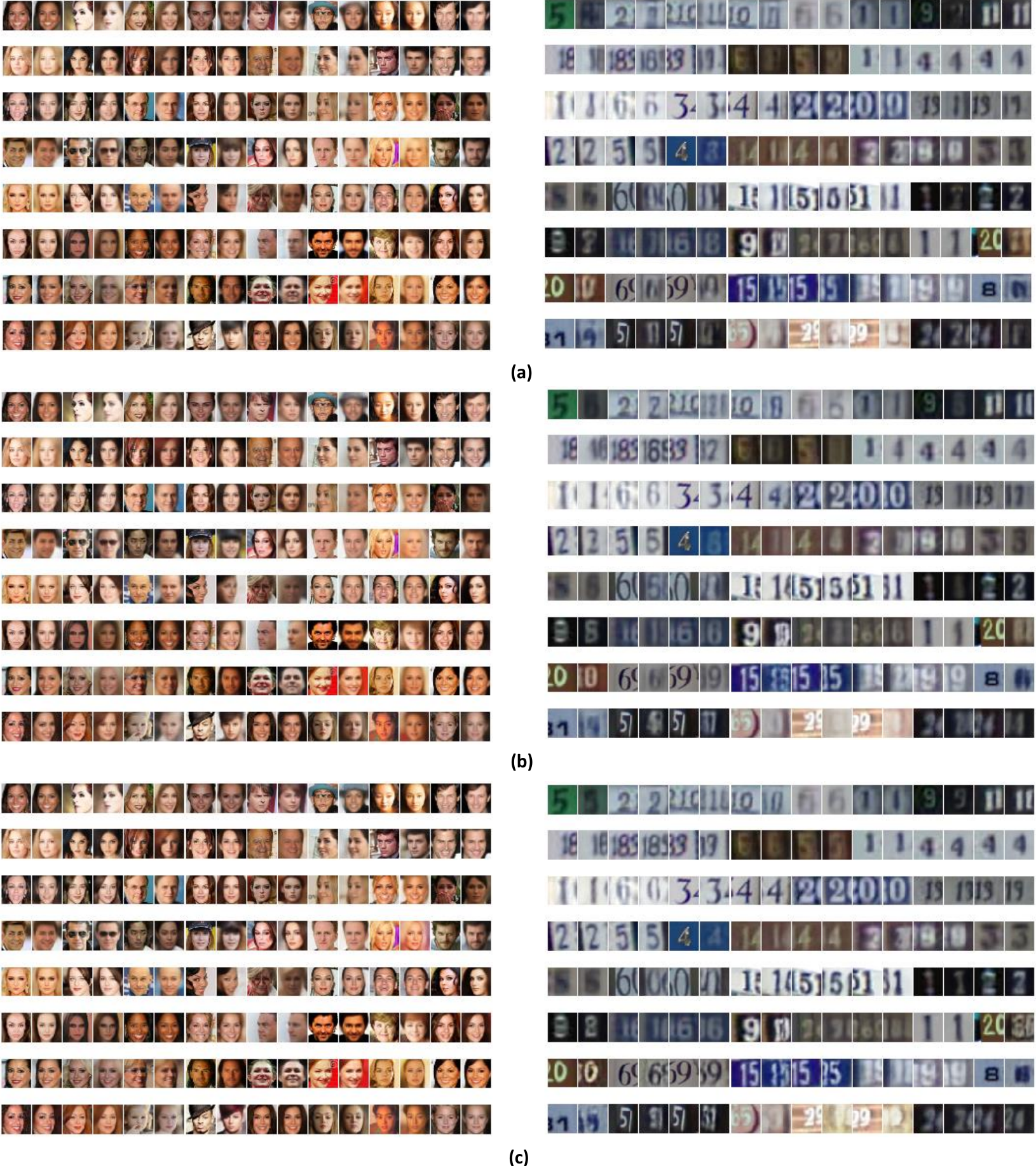}
    \caption{Reconstruction of held out data of CelebA (left) and SVHN (right) by (a) AAE \citep{makhzani2016adversarial} (b) VAE \citep{kingma2014auto} and (c) GEN (proposed). The size of the latent space for CelebA and SVHN are $l=40$ and $l=25$ respectively. Reconstruction of $64$ test images of both data sets are shown in the figure for each method, where the reconstructed image follows the test input in the grid layout.}
    \label{fig:Reconstruction}
\end{figure}

Reconstruction of the held out samples of CelebA and SVHN data set for different methods are shown in Figure \ref{fig:Reconstruction}. These reconstructed samples are produced by the corresponding models used for estimating the FID scores reported in the main paper.

%\begin{figure}[h]
%    \centering
%    \includegraphics[width=1.0\textwidth]{Figures/SVHN_Reconstructed_Final.pdf}
%    \caption{Reconstruction of held out data of SVHN by (a) AAE \citep{makhzani2016adversarial} (b) VAE \citep{kingma2014auto} and (c) GEN (proposed) in $l=25$ dimensional latent space. Reconstruction of $64$ test images are shown in the figure for each method, where the reconstructed image follows the test input in the grid.}
%    \label{fig:SVHN_Reconstruction}
%\end{figure}

%\begin{figure}[!t]
%    \centering
%    \includegraphics[width=1.0\textwidth]{Figures/CelebA_Reconstructed_Final.pdf}
%    \caption{Reconstruction of held out data of CelebA by (a) AAE \citep{makhzani2016adversarial} (b) VAE \citep{kingma2014auto} and (c) GEN (proposed) in $l=40$ dimensional latent space. Reconstruction of $64$ test images are shown in the figure for each method, where the reconstructed image follows the test input in the grid.}
%    \label{fig:CelebA_Reconstruction}
%\end{figure}

\subsection{Generated samples}
\subsubsection{Standard Normal}
\begin{figure}
    \centering
    \includegraphics[width=1.0\textwidth]{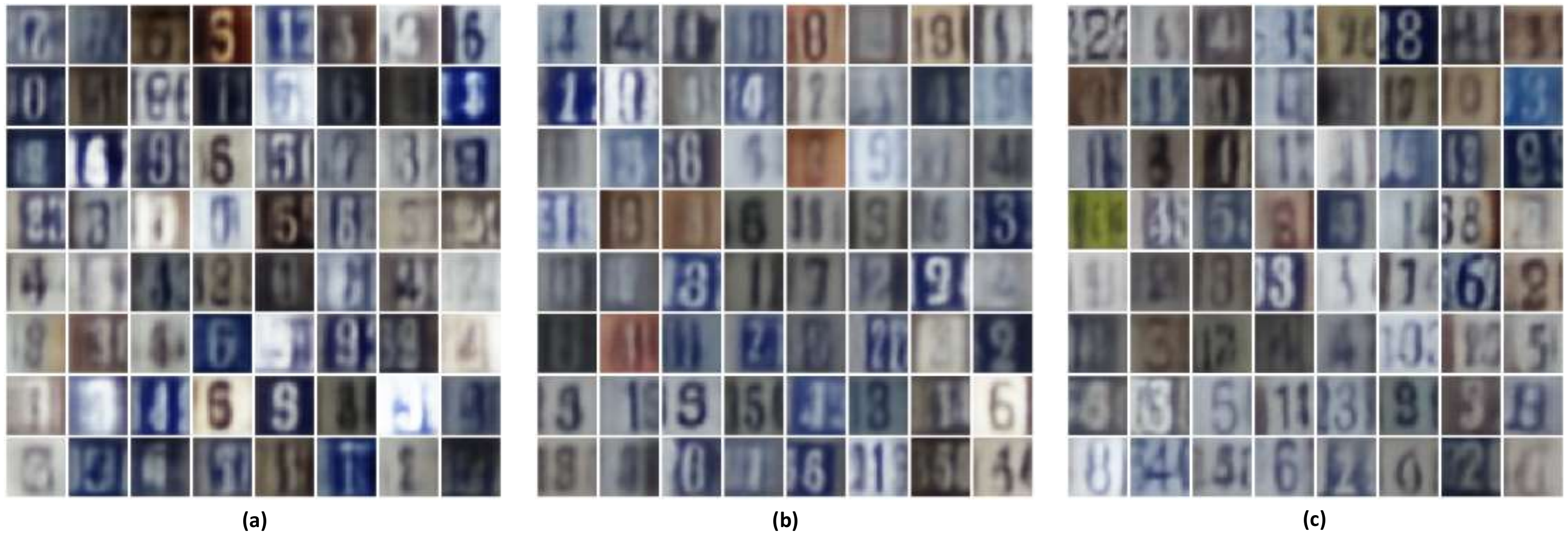}
    \caption{Samples generated by models trained with (a) AAE \citep{makhzani2016adversarial} (b) VAE \citep{kingma2014auto} and (c) GEN (proposed) over SVHN dataset in $l=25$ dimensional latent space.}
    \label{fig:SVHN_generated_samples}
\end{figure}

In Figure \ref{fig:SVHN_generated_samples}, we present the generated samples using models trained over SVHN for different methods. The target distribution is standard normal for all the methods. The generated samples are produced by the corresponding models used for estimating the FID scores.
\subsubsection{Mixture of Gaussians}
\begin{figure}
    \centering
    \includegraphics[width=1.0\textwidth]{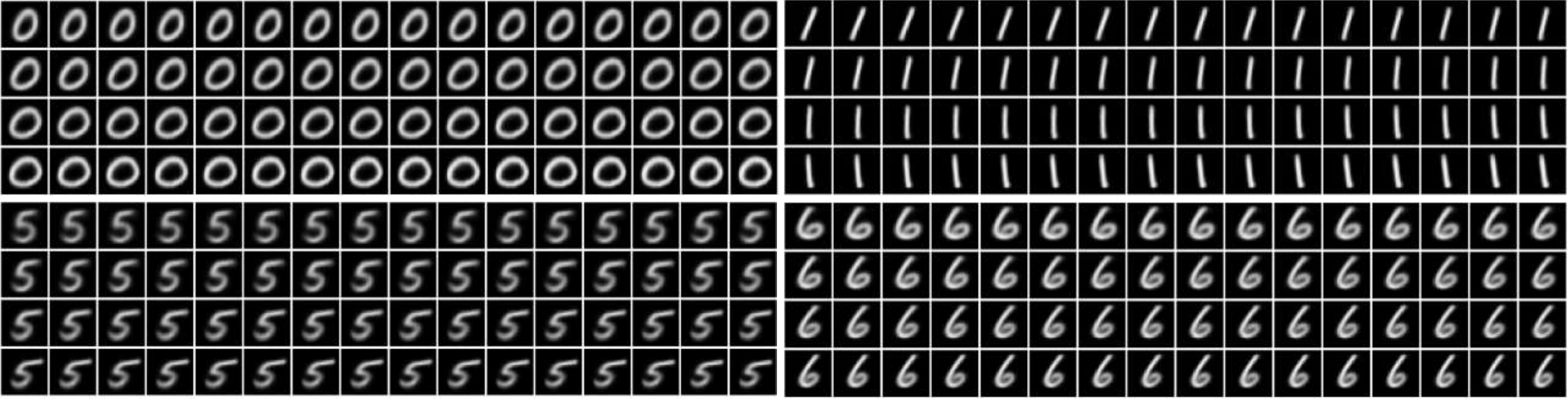}
    \caption{Samples generated by interpolation (along axes) in different modes of the mixture of Gaussians. The variation in the style of the generated samples is evident in all the modes.}
    \label{fig:MoG_Interpolation}
\end{figure}

Samples generated by interpolation in different modes of the mixture of Gaussians are reported in Figure \ref{fig:MoG_Interpolation}. Generated samples corresponds to its mode in the distribution. These results validate the matching of the distribution in the latent space and clustering of the data (based on its labels) in the semi-supervised set up.
\subsection{Novelty detection}
\begin{figure}
    \centering
    \includegraphics[width=1.0\textwidth]{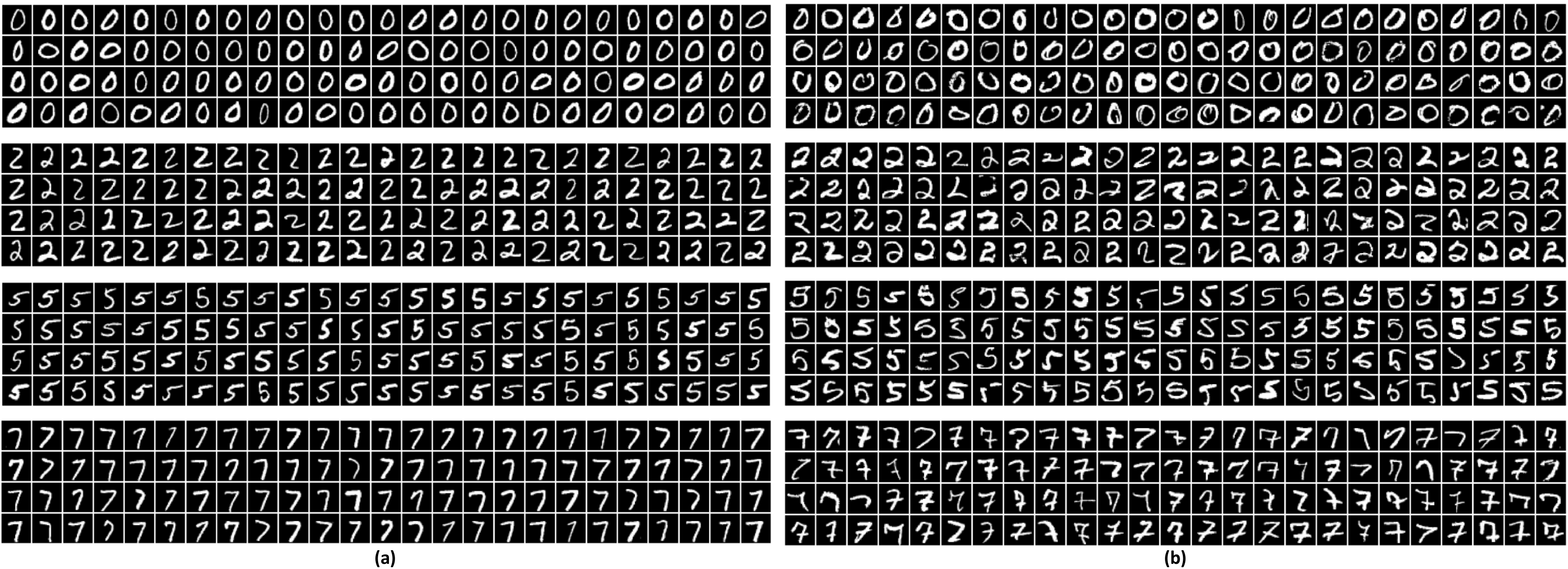}
    \caption{Novelty detection on the held out data of MNIST for digits $0, 2, 5,$ and $7$ (a) $100$ samples with low outlier score aka inliers and (b) $100$ samples having high outlier score, identified as outliers to the distribution}
    \label{fig:More_Outlier_Results}
\end{figure}

Figure \ref{fig:More_Outlier_Results} shows the performance of the proposed outlier detection method, using models trained with GENs, on the held out data of the different category of digits in the MNIST data set.

\subsection{Interpolation of the latent space}
\begin{figure}
    \centering
    \includegraphics[width=0.95\textwidth]{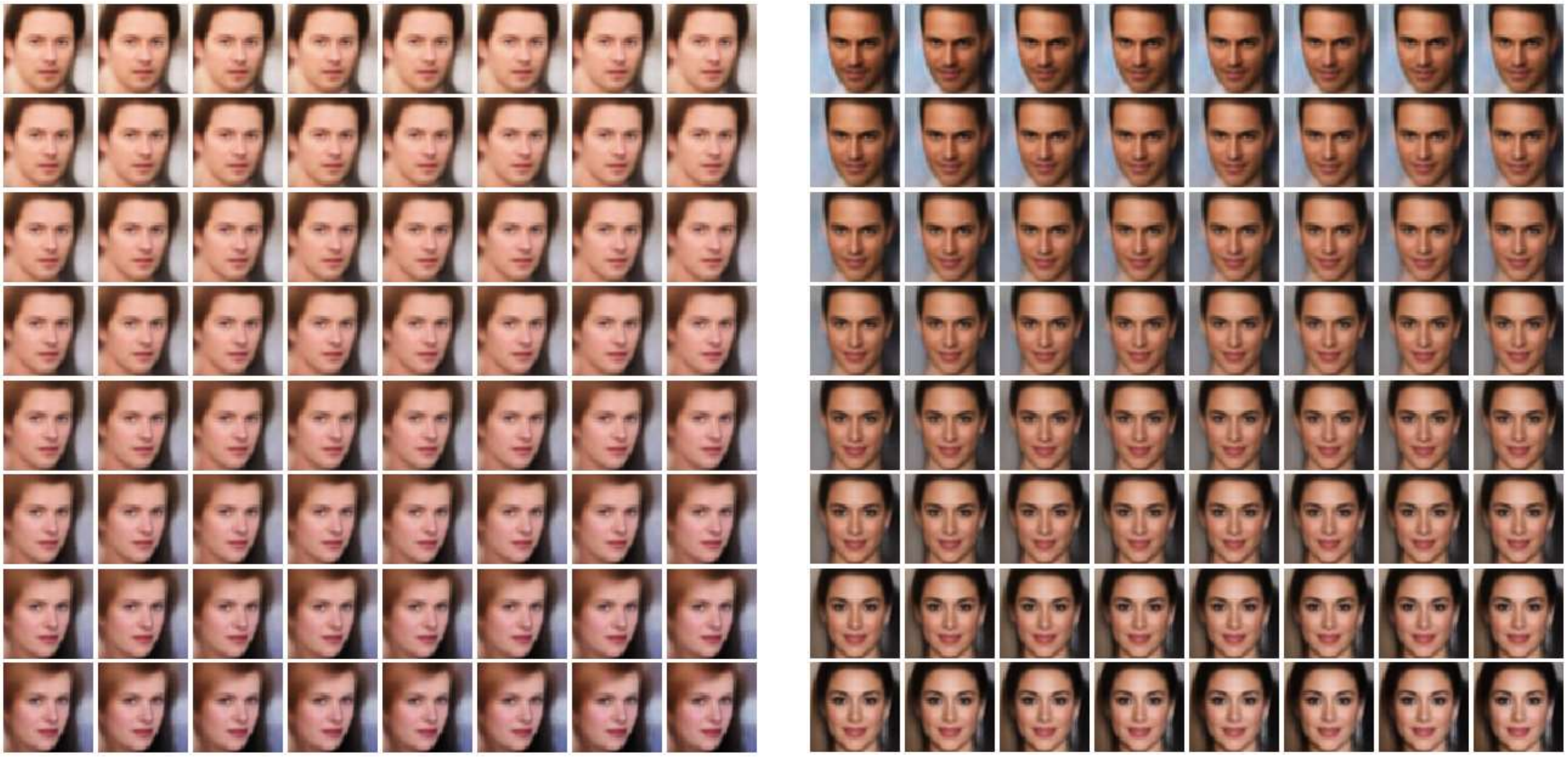}
    \caption{Images generated by interpolation between random pairs of samples in the latent space, for a model trained with GEN. The images at the left top and right bottom corners of the grid correspond to the images generated by the random pair of samples in the latent space. The interpolated results are shown across the columns in the grid, starting at the left top corner.}
    \label{fig:Interpolation_random}
\end{figure}

\begin{figure}
    \centering
    \includegraphics[width=0.95\textwidth]{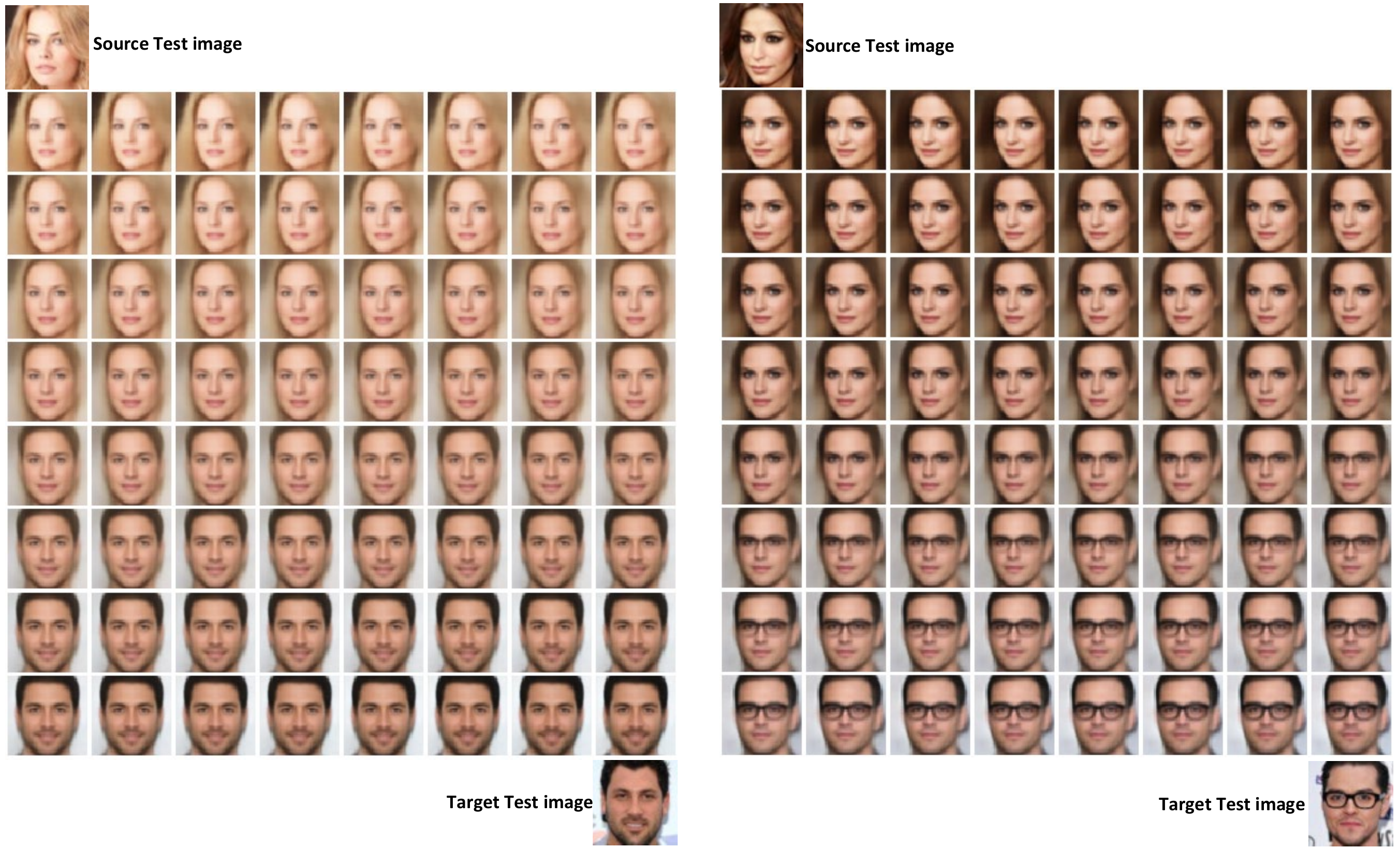}
    \caption{Images generated by interpolation between test images projected in the latent space, for a model trained with GEN. The real source and target test images are tagged along with the grid layout, showing interpolation results. We use the projection of the source and target image in the latent space, shown at the left top and right bottom corner of the grid respectively, for producing interpolation results. Images generated by interpolation are shown across the columns in the grid, starting at the left top corner.}
    \label{fig:Interpolation_test_images}
\end{figure}

In this experiment, we linearly interpolate in the latent space between random samples in the latent space and projected test samples, for a model trained with GEN. In both the tests, we found the interpolated results to generate images bearing strong facial features for any pair of latent values. Figure \ref{fig:Interpolation_random} and \ref{fig:Interpolation_test_images} show generated images by linear interpolation of the random samples in the latent space and projected test samples in the latent space, respectively.

\bibliographystyle{abbrvnat}
\setcitestyle{authoryear,open={((},close={))}}
\bibliography{supplementary_references}